\pdfoutput=1

\documentclass[11pt]{article}

\usepackage[]{acl}

\usepackage{times}
\usepackage{latexsym}
\usepackage{booktabs}
\usepackage{multirow}

\usepackage[T1]{fontenc}

\usepackage[utf8]{inputenc}

\usepackage{microtype}

\usepackage{inconsolata}

\usepackage{graphicx}

\title{IndicSQuAD: A Comprehensive Multilingual Question Answering Dataset for Indic Languages
}

  \author{Sharvi Endait$^{1,3}$, Ruturaj Ghatage$^{1,3}$, Aditya Kulkarni$^{1,3}$, Rajlaxmi Patil$^{1,3}$, Raviraj Joshi$^{2,3}$ \\
        Pune Institute of Computer Technology, Pune$^1$ \\ Indian Institute of Technology Madras, Chennai$^2$\\L3Cube Labs, Pune$^3$}

\begin{document}
\maketitle
\begin{abstract}
The rapid progress in question-answering (QA) systems has predominantly benefited high-resource languages, leaving Indic languages largely underrepresented despite their vast native speaker base. In this paper, we present IndicSQuAD, a comprehensive multi-lingual extractive QA dataset covering nine major Indic languages, systematically derived from the SQuAD dataset. Building on previous work with MahaSQuAD for Marathi, our approach adapts and extends translation techniques to maintain high linguistic fidelity and accurate answer-span alignment across diverse languages. IndicSQuAD comprises extensive training, validation, and test sets for each language, providing a robust foundation for model development. We evaluate baseline performances using language-specific monolingual BERT models and the multilingual MuRIL-BERT. The results indicate some challenges inherent in low-resource settings. Moreover, our experiments suggest potential directions for future work, including expanding to additional languages, developing domain-specific datasets, and incorporating multimodal data. The dataset and models are publicly shared at \url{https://github.com/l3cube-pune/indic-nlp}
\end{abstract}

\section{Introduction}

Question Answering (QA) has been a cornerstone of natural language understanding (NLU), with datasets like SQuAD \cite{rajpurkar2016squad} driving advancements in machine learning models for extractive QA. While these datasets have enabled significant progress, most large-scale QA datasets are centered around English and a few high-resource languages, leaving many Indic languages underrepresented. Recent benchmarks highlight this disparity, with evaluations showing that multilingual Large Language Models (LLMs) perform substantially worse on low-resource languages compared to English. As a result, developing robust QA systems for Indic languages remains a challenge, despite the large native speaker population and the growing need for AI applications across diverse linguistic landscapes.

Indic languages, including Hindi, Bengali, Tamil, Telugu, Marathi, and others, are spoken by over a billion people. However, unlike English or Chinese, Indic languages lack extensive QA datasets, limiting the performance and adaptability of multilingual models in real-world applications. While initiatives such as TyDiQA \cite{clark2020tydiqabenchmarkinformationseeking} and XQuAD \cite{Artetxe:etal:2019} have attempted to introduce non-English QA datasets, these remain either limited in size or do not comprehensively cover multiple Indic languages.
The Indic-QA Benchmark \cite{singh2025indicqabenchmarkmultilingual} introduced in 2024 represents a significant advancement in this space, covering 11 major Indian languages from two language families and incorporating both extractive and abstractive question-answering tasks. This benchmark combines existing datasets with English QA datasets translated into Indian languages, demonstrating the growing recognition of the need for comprehensive multilingual QA resources. Nevertheless, the performance of multilingual models on this benchmark remains subpar, particularly for low-resource languages, underscoring the persistent challenges in this domain. A more targeted effort in this direction is the IndicQuest benchmark \cite{rohera2024l3cubeindicquestbenchmarkquestionanswering}, which is specifically designed to evaluate the factual accuracy of Indic LLMs.

The absence of large-scale, high-quality annotated QA datasets for Indic languages restricts their integration into information retrieval, education, healthcare services, governance applications, and AI-powered customer support systems. This limitation perpetuates digital inequality, where speakers of high-resource languages benefit more from technological advancements than those speaking low-resource languages.

To bridge this gap, we introduce IndicSQuAD, a comprehensive multi-lingual QA dataset covering 9 Indic languages, built by systematically translating and adapting the English SQuAD dataset while ensuring linguistic accuracy and answer-span alignment. This work builds upon our previous research on MahaSQuAD \cite{ruturaj-etal-2023-mahasquad}, which focused on Marathi and extends the approach to nine additional Indic languages. Drawing from recent advances in alignment techniques and span retrieval methods, our approach addresses the challenges identified in previous translation efforts (e.g., morphological variations, syntactic differences, and maintaining contextual integrity). IndicSQuAD represents the largest multi-Indic QA resource to date, presenting the dataset along with additional baseline models, designed to facilitate research in low-resource language modeling and improve access to knowledge for Indic language speakers.
Our key contributions are as follows:
\begin{enumerate}
    \item \textbf{Creation of IndicSQuAD dataset\footnote{\url{https://github.com/l3cube-pune/indic-nlp/tree/main/L3Cube-IndicSQUAD}}} – A large-scale multi-lingual extractive QA dataset for 10 Indic languages, derived from SQuAD, ensuring high linguistic fidelity and comprehensive coverage across language families. The languages supported are Marathi, Hindi, Bengali, Telugu, Tamil, Gujarati, Punjabi, Kannada, Oriya, and Malayalam.
    \item \textbf{Comprehensive Baseline Models and Benchmarking} –We establish strong baseline performances using L3Cube’s in-house fine-tuned monolingual BERT models for each Indic language, ensuring a tailored evaluation that captures language-specific nuances. Additionally, we provide a comparative analysis with multilingual models like MuRIL Bert \cite{khanuja2021muril}, assessing their effectiveness across diverse Indic languages. This evaluation framework addresses the unique challenges of each language, enabling meaningful comparisons across language families and resource availability.
\end{enumerate}
\section{Related Work}

The development of question-answering systems for Indian languages has gained significant attention in recent years, though these languages remain resource-scarce compared to English. Several datasets have been created to address this gap using translation and native language approaches. This section provides a comprehensive overview of the existing work in this domain.

\subsection{Translation Based Approaches}
Our previous work, MahaSQuAD \cite{ruturaj-etal-2023-mahasquad} \cite{joshi-2022-l3cube} represents the first comprehensive question-answering dataset specifically developed for Marathi. This work filled a critical gap in the language resources landscape. The paper details that MahaSQuAD consists of 118,516 training, 11,873 validation, and 11,803 test samples, accompanied by a gold test set of 500 manually verified examples. The work also presents a generic approach for translating SQuAD into any low-resource language, addressing the significant challenge of mapping answer translations to their spans in translated passages. In the current work, we extend this approach to nine more Indian languages.

\cite{kumar2022mucot} developed an extensive resource with 28,000 samples each for Hindi and Marathi by translating SQuAD 2.0, helping address data scarcity for these languages.

XQuAD \cite{Artetxe:etal:2019} consists of 240 paragraphs and 1,190 question-answer pairs derived from SQuAD v1.1 and professionally translated into ten languages, including Hindi. This dataset has served as an important benchmark for cross-lingual question answering evaluation.

MLQA \cite{DBLP:journals/corr/abs-1910-07475} serves as another important benchmark with 4,918 context-question-answer triples available in Hindi. It enables the evaluation of cross-lingual generalization capabilities in multiple languages simultaneously.
\subsection{Natively annotated datasets}

The ChaII Dataset \cite{unknown} features context-question-answer triples in Hindi and Tamil gathered directly without translation. Created by native speaker annotators, this dataset presents realistic information-seeking tasks focused on Wikipedia articles. The dataset includes 1,104 questions with the Hindi portion translated into ten other Indian languages. 

Recent work by \cite{unknown} has investigated the application of transformer models pre-trained on multiple languages, specifically focusing on Hindi and Tamil question-answering, demonstrating enhanced performance in extractive QA tasks.

Additionally, the Extended Chaii dataset has been developed containing Tamil translations from the SQuAD dataset, designed specifically for question-answering tasks in low-resource Indic languages. The dataset consists of 2,855 training instances, 460 validation instances, and 250 test instances, making it a valuable resource for Tamil language processing.

The MMQA dataset \cite{GUPTA18.826} contains 5,495 question-answer pairs in English and Hindi, covering factoid and short descriptive questions across multiple domains. This dataset is specifically designed to evaluate both bilingual and cross-lingual question answering that processes queries in either Hindi or English and retrieves answers in either language from documents in Hindi or English. The MMQA framework represents an important contribution toward multilingual information access, particularly beneficial in the Indian context.

\subsection{Natively Annotated Datasets}
While translation-based approaches have been instrumental in creating resources for low-resource languages, natively constructed datasets offer unique advantages in preserving linguistic authenticity.

BanglaQuAD \cite{rony2024banglaquadbengaliopendomainquestion} represents a significant contribution to Bengali language processing, containing 30,808 question-answer pairs constructed directly from Bengali Wikipedia articles by native speakers. Unlike translation-based approaches, this methodology avoids potential pitfalls associated with translated datasets, including loss of linguistic authenticity and contextual accuracy. The authors provide a detailed analysis of question types and answer distributions, along with baseline performance metrics using both monolingual and multilingual models.

The INDIC QA Benchmark \cite{singh2025indicqabenchmarkmultilingual} represents one of the most recent and comprehensive efforts in this domain, covering 11 major Indian languages and addressing both extractive and abstractive QA tasks. This benchmark aims to standardize evaluation across multiple Indic languages, enabling more direct comparisons of model performance across linguistic boundaries. The benchmark incorporates various question types and difficulty levels, providing a nuanced understanding of model capabilities across different linguistic structures found in Indic languages.

L3Cube-IndicQuest \cite{rohera2024l3cubeindicquestbenchmarkquestionanswering} takes a more comprehensive approach by covering 19 Indic languages, making it one of the most linguistically diverse QA datasets available. Unlike many existing datasets that focus primarily on question-answering capabilities in general domains, L3Cube-IndicQuest specifically addresses five domains of particular relevance to the Indian context: Literature, History, Geography, Politics, and Economics. Each language subset contains carefully curated question-answer pairs designed to evaluate a model's ability to represent and process knowledge specific to Indian cultural and regional contexts.

The ChAII Dataset \cite{singh2025indicqabenchmarkmultilingual} features context-question-answer triples in Hindi and Tamil gathered directly without translation. Created by native speaker annotators, this dataset presents realistic information-seeking tasks focused on Wikipedia articles. The dataset includes 1,104 questions with the Hindi portion translated into ten other Indian languages.

\begin{table*}[t]
    \centering
    \resizebox{\textwidth}{!}{ 
    \begin{tabular}{llcccccccc}
        \toprule
       Language & Model & \multirow{2}{*}{EM\%} & \multirow{2}{*}{F1\%} & \multirow{2}{*}{EM} & \multirow{2}{*}{F1} & \multirow{2}{*}{EM} & \multirow{2}{*}{F1} & \multirow{2}{*}{BLEU\%} & \multirow{2}{*}{BLEU\%} \\\\
        & & & & (Has\_ans) & (Has\_ans) & (No\_ans) & (No\_ans) & (Unigram) & (Bigram) \\
        \midrule
        \multirow{2}{*}{Hindi} & HindiRoBERTa & 56.20 & 59.67 & 50.79 & 57.8 & 61.50 & 61.50 & 61.8 & 53.5 \\
        & MurilBERT & 53.21 & 56.89 & 53.05 & 60.49 & 53.37 & 53.37 & 62.8 & 55.0 \\
       
        \midrule
        \multirow{2}{*}{Punjabi} & PunjabiBERT & 51.04 & 54.53 & 47.59 & 54.63 & 54.42 & 54.42 & 54.4 & 46.1 \\
        & MurilBERT & 50.80 & 54.41 & 47.08 & 54.38 & 54.40 & 54.40 & 54.5 & 46.3 \\
        \midrule
        \multirow{2}{*}{Gujarati} & GujaratiBERT & 49.00 & 52.91 & 47.36 & 55.26 & 50.61 & 50.61 & 52.8 & 44.2 \\
        & MurilBERT & 48.09 & 52.24 & 47.63 & 56.00 & 48.54 & 48.54 & 54.7 & 47.0 \\
        \midrule
        \multirow{2}{*}{Kannada} & KannadaBERT & 50.97 & 54.90 & 48.54 & 56.49 & 53.34 & 53.34 & 52.3 & 45.9 \\
        & MurilBERT & 49.64 & 53.81 & 48.27 & 56.68 & 50.99 & 50.99 & 53.7 & 44.6 \\
        \midrule
        \multirow{2}{*}{Tamil} & TamilBERT & 50.97 & 54.44 & 46.38 & 53.39 & 55.47 & 55.47 & 53.03 & 44.26 \\
        & MurilBERT & 49.70 & 53.14 & 46.40 & 53.34 & 52.94 & 52.94 & 53.85 & 44.83 \\
        \midrule
        \multirow{2}{*}{Bengali} & BengaliBERT & 50.07 & 54.27 & 46.93 & 55.42 & 53.14 & 53.14 & 57.7 & 49.5 \\
        & MurilBERT & 49.36 & 53.70 & 46.98 & 55.75 & 51.70 & 51.70 & 56.6 & 48.2 \\
        \midrule
        \multirow{2}{*}{Telugu} & TeluguBERT & 52.17 & 55.34 & 44.98 & 51.37 & 59.24 & 59.24 & 54.8 & 47.5 \\
        & MurilBERT & 51.11 & 54.38 & 44.30 & 50.90 & 57.82 & 57.82 & 52.9 & 45.1 \\
        \midrule
        \multirow{2}{*}{Oriya} & OdiaBERT & 54.33 & 57.60 & 44.61 & 51.22 & 63.82 & 63.82 & 56.8 & 48.6 \\
        & MurilBERT & 48.65 & 52.47 & 43.75 & 51.48 & 53.44 & 53.44 & 49.7 & 41.7 \\
        \midrule
        \multirow{2}{*}{Malayalam} & MalayalamBERT & 51.02 & 49.42 & 42.24 & 49.26 & 59.59 & 59.59 & 52.0 & 43.2 \\
        & MurilBERT & 45.67 & 49.62 & 41.89 & 49.89 & 49.36 & 49.36 & 46.9 & 38.6 \\
        \midrule
        \multirow{2}{*}{Marathi} & MahaBERT & 51.28 & 54.88 & 51.04 & 58.31 & 51.52 & 51.52 & 57.9 & 49.9 \\
        & MurilBERT & 50.13 & 53.91 & 51.26 & 58.92& 49.03 & 49.03 & 57.7 & 49.4 \\
        \bottomrule
    \end{tabular}
    }
    \caption{Performance of various models on different languages}
    \label{tab:performance}
\end{table*}
\section{Experimental Setup}

\subsection{Data Collection}
Our data collection process utilized the Stanford Question Answering Dataset (SQuAD 2.0), originally in English, which comprises over 150,000 question-answer pairs. Notably, about 34\% of these questions are unanswerable, challenging models to handle ambiguity and non-definitive answers effectively. Each row in SQuAD 2.0 includes essential components such as title, context, question, answer start index, and answer text.

To create datasets, we emulated the robust translation and transliteration procedure used for the Marathi language, in creating MahaSQuAD. This involved developing a sophisticated algorithm to address inconsistencies that arise during translation, particularly in locating the correct answer index post-translation. By doing so, we ensured that the translated datasets accurately reflect the nuances of the original English dataset while adapting to the linguistic characteristics of the target language. This approach not only enhances the quality of the translated datasets but also facilitates the development of more accurate question-answering models for low-resource languages.

\subsubsection{Translation Methodologies}
When creating multilingual QA datasets through translation, several methodologies can be employed, each with its advantages and challenges. Beyond our approach based on MahaSQuAD \cite{ruturaj-etal-2023-mahasquad}, other researchers have explored various techniques for cross-lingual transfer in QA contexts.

\cite{kumar2022mucotmultilingualcontrastivetraining} proposed Multilingual Contrastive Training (MuCoT), a three-stage pipeline for question-answering in low-resource languages. This approach utilizes translation and transliteration with contrastive training across language families, showing particular effectiveness when data from the same language family is grouped. Their experiments demonstrated that translations from Indo-Aryan languages (Bengali and Marathi) significantly improved performance on Hindi, while Dravidian language data (Telugu and Malayalam) enhanced Tamil performance.

More recently, Self-Translate-Train \cite{ri2024selftranslatetrainenhancingcrosslingualtransfer} has emerged as a promising approach that leverages large language models to generate translations without requiring external translation systems. This method generates synthetic training data in the target language by utilizing the model's own translation capabilities, demonstrating substantial performance gains across several non-English languages without intensive additional data collection.

In creating IndicSQuAD, we built upon these approaches while addressing the specific challenges of Indic languages, such as maintaining context and handling linguistic nuances during translation. Our methodology focused particularly on ensuring accurate mapping of answer spans in translated passages, a significant challenge when dealing with languages that differ substantially in word order and sentence structure from English.

\begin{table}[h]
    \centering
    \begin{tabular}{lll}
        \toprule
        \textbf{Language} & \textbf{Family} & \textbf{Script} \\
        \midrule
        Marathi   & Indo-Aryan  & Devanagari \\
        Hindi     & Indo-Aryan  & Devanagari \\
        Punjabi   & Indo-Aryan  & Gurmukhi   \\
        Bengali   & Indo-Aryan  & Bengali    \\
        Gujarati  & Indo-Aryan  & Gujarati   \\
        Oriya      & Indo-Aryan  & Oriya       \\
        Tamil     & Dravidian   & Tamil      \\
        Telugu    & Dravidian   & Telugu     \\
        Kannada   & Dravidian   & Kannada    \\
        Malayalam & Dravidian   & Malayalam  \\
        \bottomrule
    \end{tabular}
    \caption{Languages, their Families, and Scripts}
    \label{tab:languages}
\end{table}

\subsection{Languages covered}
IndicSQuAD includes question-answering datasets for 9 Indic languages, covering a diverse set of Indo-Aryan and Dravidian languages. These languages vary significantly in terms of script, morphology, and linguistic resources, making the dataset a valuable resource for multilingual and low-resource NLP research. 

India's linguistic diversity is immense, with the 2011 Census identifying 122 major languages and 1,599 other languages. Among these, the most widely spoken languages are Hindi, Bengali, Marathi, Telugu, Tamil, Gujarati, Punjabi, Kannada, Odia, and Malayalam. Despite their extensive use, many of these languages are considered low-resource in the field of Natural Language Processing (NLP) due to the limited availability of annotated datasets and linguistic tools. This scarcity poses significant challenges in developing robust NLP applications, as models trained on high-resource languages often fail to generalize effectively to low-resource contexts. By creating comprehensive question-answering datasets for these languages, IndicSQuAD aims to bridge this gap, facilitating the development of more inclusive and effective NLP applications.

\subsection{Robust approach}
The creation of IndicSQuAD employed a robust translation strategy to preserve linguistic accuracy and contextual integrity in low-resource languages. To address the challenge of aligning translated answers with their corresponding spans in translated passages, the English context was first segmented into sentences. Each sentence and its associated answer were then translated into the target language. Using similarity analysis tools, the most contextually appropriate span in the translated passage was identified to match the translated answer. This approach ensured precise alignment, overcoming the common mismatch between independently translated answers and contexts.

\begin{figure}[t]
\centering
    \includegraphics[width= 5 cm]{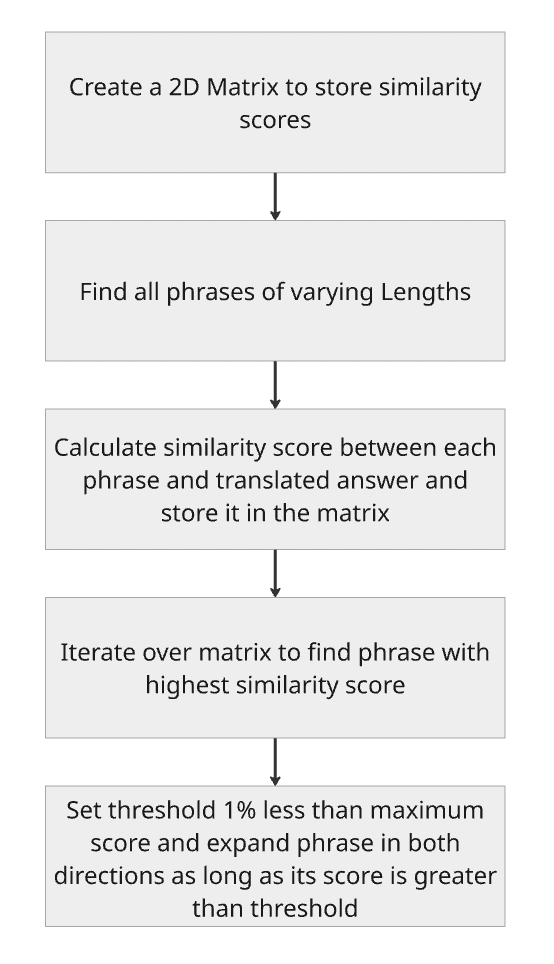}
    \caption{Algorithm for obtaining the answer and the answer span from the context}
    \label{fig:sentenceprocessing}
\end{figure}
Figure 1 illustrates the methodology employed to accurately map translated answers to their corresponding spans within translated passages. The robust algorithm developed for MahaSQuAD ensures precise alignment of translated answers within their contexts through the following steps: 
\begin{enumerate}
    \item \textbf{Sentence Segmentation}: The English context is divided into individual sentences using the NLTK library.
    \item \textbf{Answer Sentence Identification}: The English sentence containing the answer is identified from the individual sentences.
    \item \textbf{Translation}: Both the identified sentence and the answer are translated into Marathi (target language) using Google Translate.
    \item \textbf{Similarity Analysis}: Within the translated Marathi sentence, all possible substrings are compared to the translated answer using the SimilarityAnalyzer from the MahaNLP library \cite{joshi2022l3cube-mahanlp,magdum2023mahanlp}. The library uses embedding models released in \cite{deode2023l3cube}. A similarity score matrix is generated to identify the substring with the highest similarity to the translated answer.
    \item \textbf{Answer Span Determination}: The substring with the maximum similarity is selected as the base answer. Adjacent words are appended to this base answer, and the similarity is recalculated. If the new similarity score remains within 1\% of the maximum, the extended phrase is accepted. This iterative process ensures that the translated answer accurately reflects the original meaning and context.
    \item \textbf{Transliteration}: To maintain script consistency, named entities and numerical values are transliterated into the Devanagari script using the AI4Bharat Transliteration Engine.
\end{enumerate}

To further enhance consistency, named entities were transliterated into Devanagari script using the AI4Bharat Transliteration Engine, and numerical values were converted into their counterparts. This meticulous process minimized errors and ensured uniformity across the dataset. The resulting dataset, comprising 118,516 training samples, 11,873 validation samples, and 11,803 test samples, provides a scalable framework for translating SQuAD into other low-resource languages while maintaining linguistic and cultural nuances.

\subsection{Dataset Statistics}
Each of the ten languages in IndicSQuAD, including Marathi, Hindi, Bengali, Telugu, Tamil, Gujarati, Punjabi, Kannada, Oriya, and Malayalam, consists of \textbf{118,516 entities} in the training set, \textbf{11,873 entities} in the validation set, and \textbf{11,803 entities} in the test set. This large-scale dataset provides a robust foundation for training and evaluating QA models, addressing the scarcity of high-quality annotated resources for Indic low-resource languages.

\begin{table}[h]
    \centering
    \begin{tabular}{|c|c|}
    \hline
         Training set & 118,516 samples \\
         Validation set & 11,873 samples \\
         Test set & 11,803 samples\\
         \hline
    \end{tabular}
    \caption{Dataset Statistics}
    \label{tab:my_label}
\end{table}

\section{Benchmarking and Experiments}

\subsection{Models used}
To evaluate IndicSQuAD, we employed monolingual and multilingual models to establish baseline performances across the ten Indic languages.
\begin{itemize}
    \item \textbf{Monolingual Models} \\Monolingual models are language-specific models trained solely on a particular language. These models are optimized for the linguistic characteristics of their respective languages, often leading to improved performance compared to multilingual counterparts. For each language in IndicSQuAD, we utilized L3Cube’s BERT-based monolingual models low resource languages such as HindiBERT \cite{joshi2022l3cubehind}, PunjabiBERT, GujaratiBERT, KannadaBERT, TamilBERT, BengaliBERT, OdiaBERT, and MalayalamBERT, fine-tuned on the corresponding datasets.
   \item \textbf{Multilingual Models} \\In our research, we utilized MuRIL-BERT (Multilingual Representations for Indian Languages), a transformer-based language model pre-trained on 17 Indian languages, including Marathi. MuRIL-BERT has demonstrated robust performance in understanding and processing Indian languages, making it a suitable choice for our study. This model's architecture allows it to effectively capture linguistic nuances across multiple Indian languages, facilitating the development of more accurate and efficient natural language processing applications.
\end{itemize}

\subsection{Experimental Setup}
We conducted fine-tuning on our models using a custom dataset spanning three epochs and utilizing A100 GPUs with a consistent batch size of 32. The carefully selected hyperparameters include n\_best\_size ( which refers to the number of predictions provided per question ) set to 2, which significantly shaped the training dynamics and influenced the experimental outcomes. The other key hyperparameters employed during fine-tuning included a learning rate of 1e-4 and the AdamW optimizer. These adjustments were crucial in refining the model and enhancing its performance.

\subsection{Results}

The evaluation of the IndicSQuAD dataset from Table 1 highlights the superior performance of monolingual models over the multilingual MurilBERT across most Indic languages. For example, HindiRoBERTa outperformed MurilBERT for Hindi, achieving higher EM and F1 scores (56.20\% and 59.67\%, respectively, compared to 53.21\% and 56.89\%). Similarly, language-specific models like BengaliBERT and TamilBERT demonstrated better results in their respective languages, with BengaliBERT achieving an EM score of 50.07\% compared to MurilBERT's 49.36\%. These monolingual models consistently showed better contextual understanding and exact match accuracy.

In contrast, MurilBERT exhibited more generalized performance but lagged in capturing language-specific nuances, especially in low-resource languages like Telugu and Malayalam. For instance, TeluguBERT achieved an EM score of 52.17\%, outperforming MurilBERT's 51.11\% by leveraging its tailored design for Telugu. This trend underscores the importance of monolingual models in improving language-specific performance and highlights the need for further optimization of multilingual models to close the gap in low-resource language processing.

\section{Conclusion and Future work}
IndicSQuAD represents a significant advancement in addressing the scarcity of high-quality, large-scale question answering datasets for Indic languages. By systematically translating and adapting the widely-used SQuAD dataset into nine major Indic languages, this work not only bridges the resource gap but also establishes robust baselines using both language-specific and multilingual models. The comprehensive evaluation framework highlights the superior performance of monolingual models in capturing linguistic nuances, while also underscoring the challenges faced by multilingual models in low-resource settings. The dataset, along with the accompanying models and evaluation tools, is publicly available, fostering further research and development in multilingual NLP. Moving forward, expanding IndicSQuAD to additional languages, creating domain-specific datasets, and integrating multimodal data will further enhance the accessibility and effectiveness of AI-powered applications for Indic language speakers. This initiative is a crucial step toward reducing digital inequality and ensuring that speakers of low-resource languages can fully benefit from advances in natural language understanding and information retrieval.\\
While IndicSQuAD provides a strong foundation for question-answering (QA) tasks in Indic languages, there are several directions for future research and development:
\begin{enumerate}
    \item \textbf{Expansion to More Languages} \\Extending the dataset to cover additional low-resource Indic languages, such as Assamese, Manipuri, and Santali, to improve multilingual accessibility and representation.
    \item \textbf{Domain-Specific QA Datasets} \\ Creating specialized datasets for legal, medical, and financial domains to improve real-world applicability in Indic languages.
    \item \textbf{Multimodal QA for Indic Languages} \\ Extending the dataset to incorporate images, videos, and speech, enabling multimodal question-answering for a more inclusive AI ecosystem.
    \item \textbf{Interactive and Real-World Applications} \\ Deploying QA models trained on IndicSQuAD into real-world applications, such as chatbots, voice assistants, and educational tools, to enhance accessibility and usability.
\end{enumerate}

\section*{Acknowledgments}

This work was done under the L3Cube Pune mentorship program. We want to thank our mentors at L3Cube for their continuous support and encouragement. This work is a part of the L3Cube-IndicNLP project.


\bibliography{main}

\appendix
\section{Appendix}

\begin{table}[h]
\centering
\begin{tabular}{|l|l|}
\hline
\textbf{Language} & \textbf{Model link} \\
\hline
Marathi & \href{https://huggingface.co/l3cube-pune/marathi-question-answering-squad-bert}{marathi-question-answering-squad-bert} \\
Hindi & \href{https://huggingface.co/l3cube-pune/hindi-question-answering-squad-bert}{hindi-question-answering-squad-bert} \\
Bengali & \href{https://huggingface.co/l3cube-pune/bengali-question-answering-squad-bert}{bengali-question-answering-squad-bert} \\
Telugu & \href{https://huggingface.co/l3cube-pune/telugu-question-answering-squad-bert}{telugu-question-answering-squad-bert} \\
Tamil & \href{https://huggingface.co/l3cube-pune/tamil-question-answering-squad-bert}{tamil-question-answering-squad-bert} \\
Gujarati & \href{https://huggingface.co/l3cube-pune/gujarati-question-answering-squad-bert}{gujarati-question-answering-squad-bert} \\
Punjabi & \href{https://huggingface.co/l3cube-pune/punjabi-question-answering-squad-bert}{punjabi-question-answering-squad-bert} \\
Kannada & \href{https://huggingface.co/l3cube-pune/kannada-question-answering-squad-bert}{kannada-question-answering-squad-bert} \\
Oriya & \href{https://huggingface.co/l3cube-pune/oriya-question-answering-squad-bert}{oriya-question-answering-squad-bert} \\
Malayalam & \href{https://huggingface.co/l3cube-pune/malayalam-question-answering-squad-bert}{malayalam-question-answering-squad-bert} \\
\hline
\end{tabular}
\caption{Language-specific SQUAD BERT models on HuggingFace}
\end{table}


\end{document}